\documentclass{article}

\usepackage[utf8]{inputenc} 
\usepackage[T1]{fontenc}    
\usepackage{hyperref}       
\usepackage{url}            
\usepackage{booktabs}       
\usepackage{amsfonts}       
\usepackage{nicefrac}       
\usepackage{microtype}      
\usepackage{times}
\usepackage{graphicx,xspace} 
\usepackage[dvipsnames]{xcolor}
\usepackage{epstopdf}
\usepackage{subfigure}
\usepackage{amsmath,amssymb,amsthm,array,amsfonts,mathtools}

\usepackage{algorithm}
\usepackage{algorithmic}

\usepackage{booktabs,multirow,rotating}
\usepackage{hyperref}

\usepackage{wrapfig}

\DeclareGraphicsExtensions{.pdf,.jpeg,.png,.jpg}

\makeatletter
\DeclareRobustCommand\onedot{\futurelet\@let@token\@onedot}
\def\@onedot{\ifx\@let@token.\else.\null\fi\xspace}

\def\eg{\emph{e.g}\onedot} 
\def\ie{\emph{i.e}\onedot}

\makeatother

\def\Vec#1{{\boldsymbol{#1}}}
\def\Mat#1{{\boldsymbol{#1}}}

\newcommand{\eat}[1]{}

\usepackage[preprint]{nips_2018}

\title{Controllable Image-to-Video Translation:\\ A Case Study on Facial Expression Generation}

\author{
  Lijie Fan\\
  MIT CSAIL
  \And
  Wenbing Huang\\
  Tencent AI Lab
  \And
  Chuang Gan\\
  MIT-Waston Lab
  \And
  Junzhou Huang\\
  Tencent AI Lab
  \And
  Boqing Gong\\
  Tencent AI Lab
}

\begin{document}

\maketitle

\begin{abstract}
The recent advances in deep learning have made it possible to generate photo-realistic images by using neural networks and even to extrapolate video frames from an input video clip. In this paper, for the sake of both furthering this exploration and our own interest in a realistic application, we study image-to-video translation and particularly focus on the videos of facial expressions. This problem challenges the deep neural networks by another temporal dimension comparing to the image-to-image translation. Moreover, its single input image fails most existing video generation methods that rely on recurrent models. We propose a user-controllable approach so as to generate video clips of various lengths from a single face image. The lengths and types of the expressions are controlled by users. To this end, we design a novel neural network architecture that can incorporate the user input into its skip connections and propose several improvements to the adversarial training method for the neural network. Experiments and user studies verify the effectiveness of our approach. Especially, we would like to highlight that even for the face images in the wild (downloaded from the Web and the authors' own photos), our model can generate high-quality facial expression videos of which about 50\% are labeled as real by Amazon Mechanical Turk workers. Please see our project page\footnote{\url{http://lijiefan.me/project_webpage/image2video_arxiv/index.html}} for more demonstrations.
\end{abstract}

\section{Introduction}
Upon observing the accomplishments  of deep neural networks in a variety of subfields of AI, researchers have gained keen interests in pushing its boundaries forward. Among the new domains in which they have recently achieved remarkable results, photo-realistic image generation~\cite{goodfellow2014generative,karras2017progressive} and image-to-image translation~\cite{isola2016image,zhu2017unpaired} are two well-known examples --- they were considered very difficult in general as the desired output is extremely high-dimensional, incurring the curse of dimensionality to conventional generative models. In this paper, for the sake of both furthering this exploration and our interest in a realistic application, we propose to study image-to-video translation  which challenges the deep models by yet another temporal dimension. We focus on a special case study: how to generate video clips of rich facial expressions from a single profile photo of the neutral expression.

The image-to-video translation might seem like an ill-posed problem because the output has much more unknowns to fill in than the input values. Although there have been some works on video generation~\cite{oh2015action,mathieu2015deep,lotter2016deep,villegas2017learning,tulyakov2017mocogan,liu2017video,vondrick2017generating}, they usually take as input multiple video frames and then extrapolate the future from the recurrent pattern inferred from the input, preventing them from tackling the image-to-video translation whose input supplies no temporal cue at all. Moreover, it is especially difficult to generate satisfying video clips of facial expressions for the following two reasons. One is that humans are familiar with and sensitive about the facial expressions. Any artifacts, no matter in the spatial dimensions or along the temporal dimension, could be noticed by users. The other is that the face identity is supposed to be preserved in the generated video clips. In other words, the neural network cannot remember the faces seen in the training stage but instead learn the ``imagination'' capabilities so as to handle new faces in the deployment stage.

Despite the difficulties discussed above, we believe it is feasible to tackle the image-to-video translation at least in the particular domain of facial expression generation. First, different people express emotions in similar manners. For instance, one often opens its mouth when s/he becomes excited or surprised.  Second, the expressions are often ``unimodal'' for a fixed type of emotion. In other words, there exists a procedure of gradual change from the neutral mode to the peak state of an expression. For instance, one increases her/his degree of happiness monotonically until s/he reaches the largest degree of expression. Third, the human face of a profile photo draws a majority of users' attention, leaving the quality of the generated background less important.  All these characteristics significantly reduce the variability of the video frames, making the image-to-video translation plausible.

In this paper, we propose a user-controllable approach to the image-to-video translation. Given a single profile photo as input and a target expression (e.g., happiness), our model generates several video clips of various lengths. We allow users to conveniently control the length of a video clip by specifying an array of real numbers between 0 and 1. Each number indicates the expression degree (e.g., 0.6 out of 1) the corresponding frame is supposed to depict. Moreover, our approach can generate a video frame of a particular degree  of laughing, for example, without the need of rendering the frames before it. In contrast, most existing video generation methods~\cite{oh2015action,mathieu2015deep,lotter2016deep,villegas2017learning,tulyakov2017mocogan,liu2017video,vondrick2017generating} cannot due to their recurrent generators.  Two notable exceptions are~\cite{xue2016visual} and~\cite{haocontrollable}. However, their goals differ from ours; the former predicts the probabilistic future of the input while the latter takes as input both a video frame and sparse trajectories.

We design our deep neural network and the training losses in the following manner in order to achieve the aforementioned properties. The frame generator consists of three modules: a base encoder, a residual encoder, and a decoder taking as input from both encoders. We weigh the skip connections between the residual encoder and the decoder using the expression degrees supplied by users in the test stage. In the training stage, we infer the degrees by assigning 0 to the neutral expression frame, 1 to the frame of the peak expression, and then numbers between 0 and 1 to the frames in between in proportion to their distances to the neutral frame. We train our model following the practice of generative adversarial nets~\cite{goodfellow2014generative} with the following improvements. Noting the importance of the mouth region in expressing emotions, we use a separate discriminator to take care of it. Besides, we regularize the change between adjacent frames to ensure smoothness along the temporal dimension. Finally, we augment the main task of frame generation by predicting the face landmarks.

Extensive experiments and user studies verify  that the video clips generated by our approach are of superior quality over those by the competing methods. We would like to highlight that, by even inputting the face images in the wild (downloaded from the Web and the authors' own photos), our model can generate almost realistic facial expression videos, of which around 50\% are labeled as real by Amazon Mechanical Turk workers.

\section{Related Work}
\textbf{Image-to-Image Translation.}
Image-to-image translation has re-gained much attention due to the recent advances of deep generative models~\cite{goodfellow2014generative}.
Earlier, researchers usually formulate this task as per-pixel classification or regression~\cite{long2015fully}, where the training loss conditioning on the input image is applied to each pixel such as conditional random fields~\cite{chen2018deeplab} and nonparametric loss~\cite{li2016combining}. More recent approaches apply the conditional GAN as a structured loss to penalize the joint configuration of the output, such as the Pixe2Pixel framework by~\cite{isola2017image}. Subsequently, the translation between two unpaired domains is also studied as CycleGAN~\cite{zhu2017unpaired} and the unsupervised domain adaption method in~\cite{liu2017unsupervised}. Comparing with them,  our image-to-video task is  more challenging because the temporal dynamics have to be captured in our task.

\textbf{Video Generation.} Predicting the future may benefit many applications, such as learning feature representations~\cite{goroshin2015learning,ranzato2014video,srivastava2015unsupervised} and interactions~\cite{finn2016unsupervised}.
Previous works on video generation can be roughly divided into two categories: unconditional video generation and video prediction.
The first focuses mainly generates short video clips from random vectors sampled from a prior distribution~\cite{vondrick2016generating,tulyakov2017mocogan}. VGAN~\cite{vondrick2016generating} does this by separately generating the static background and the foreground. MoCoGAN~\cite{tulyakov2017mocogan} decomposes the motion and content into two subspaces where the motion trajectory is learned by a Recurrent Neural Network (RNN). The second category, i.e., video prediction, aims at extrapolating or interpolating video frames from the observed frames~\cite{oh2015action,mathieu2015deep,lotter2016deep,villegas2017learning}.
Early work focuses on small patches~\cite{sutskever2009recurrent}. Owing to the development of deep learning, recent approaches in video prediction have shifted from predicting
patches to full frame prediction~\cite{oh2015action}. For example, \cite{mathieu2015deep} proposed an adversarial loss for video prediction and a multi-scale network architecture that results in high quality prediction for a few time steps in natural video.
Upon observing that the frame prediction quality by~\cite{mathieu2015deep} degrades quickly, the HP method by~\cite{villegas2017learning} generates the long-term feature frames by first learning the evolution of the high-level structure (\eg the pose) with a RNN and then constructing the current image frame conditioned on the predicted high-level structure and a image in the pass. A more recent work by~\cite{haocontrollable} attempts to control the video prediction by using user-defined sparse trajectories.
Our image-to-video translation is in the same vein as the video prediction, but we emphasize some of its unique characteristics. First, our task requires one single input image other than multiple video frames, opening the door for more potential applications. Second, unlike~\cite{oh2015action,mathieu2015deep,lotter2016deep,villegas2017learning} where recurrent models are applied, our method can skip an arbitrary number of frames during inference and training.

\textbf{Facial Attribute Manipulation.}
Several works~\cite{shen2017learning,lu2017conditional} have been conducted for facial images manipulation. The study by~\cite{shen2017learning} addresses the face attribute manipulation by modifying a face image according to attributes. The approach by~\cite{lu2017conditional} performs attribute-guided face image generation on unpaired image data. Since both of the above methods are mainly for  static face generation, they are not naturally applicable for our task to generate continues videos of facial expressions.

\section{Approach}
We first formalize the image-to-video translation problem and then describe our approach in detail.
\vspace{-7pt}

\subsection{Problem formulation}
Given an input image $\Mat{I}\in\mathbb{R}^{H\times W\times 3}$ where $H$ and $W$ are respectively the height and width of the image, our goal is to generate a sequence of video frames $\{\Mat{V}(a):=f(\Mat{I},a); a\in[0,1]\}$, where $f(\Mat{I},a)$ denotes the model to be learned. Note that the variable $a$, called an action variable, takes continuous values between 0 and 1, implying that there could be an infinite number of frames in the generated video clip. In practice, we allow users to give an arbitrary number of values to  $a$ and, for each of them, our model generates a frame. For simplicity, we use a separate model $f(\Mat{I},a)$ for each type of facial expressions to describe our approach.

We demand the following properties from the model $f(\Mat{I},a)$. It is supposed to reconstruct the input image when $a=0$, \ie, $f(\Mat{I}, 0)=\Mat{I}$. Besides, the function $f(\Mat{I},a)$ has to be smooth with respect to the input $a$. In other words, the generated video frames $\Mat{V}(a)$ and $\Mat{V}(a+\Delta a)$ should be visually similar when $\Delta a$ is small. The larger $a$ is, the bigger change the generated frame $\Mat{V}(a)$ is from the original image $\Mat{I}$. In the case of facial expression generation, we let $\Mat{V}(1)$ be the peak state of the expression (e.g., the state when one's mouth opens to the most when s/he laughs).

\eat{
The frames of a video beginning with $\Mat{I}_0$ by controlling the value of the action variate.
Formally, the generation process is given by
\begin{eqnarray}
\label{Eq:problem}
\Mat{V}(t) &=& f(\Mat{I}_0, a(t)),\quad a(t)\in[0,1],
\end{eqnarray}
where $\Mat{V}(t)$ is the output frame at time $t$, $a(t)$ is the continuous action variate function, and $f$ is the target function we want to learn.
Specifically, $f$ satisfies the following properties. \textbf{I.} It can reconstruct the input image initially, \ie, $f(\Mat{I}_0, 0)=\Mat{I}_0$, and converges to the stationary state of the action when $a(t)=1$. \textbf{II.} The function increases monotonically with respect to the action $a(t)$, a larger value of $a(t)$ corresponded to a stronger change from the initial state. Note that our video generation problem degenerates to the image-to-image translation task if the action variate is always selected as 1.
}

The way we formalize the frame generator $f(\Mat{I},a)$ implies several advantages over the popular recurrent models for video generation. First, the generation process is controllable. One may control the total number of frames by supplying the proper number of values for the action variable $a$. One may also tune the position of the peak state of the expression in the video. For instance, an array of monotonically increasing values let the subject of the input image express his emotion from mild to strong, while a unimodal array like $\boldsymbol{a}=[0,0.1,\cdots,1,0.9,\cdots,0]$ makes the subject express to the most and then cool down. Besides, the frames to be generated are independent of each other, taxing less over the format of the training data --- temporal smoothness is enforced by a regularization term. Finally, this model structure also benefits the optimization procedure because we do not need backpropagate gradients through time, avoiding the potential caveat of vanishing gradients.

\eat{
\textbf{Action variate.}
Training our model requires to input $a(t)$ for each frame $\Mat{V}(t)$. In this paper, all video are assumed to be aperiodic and contain single complete action. We compute the value of $a(t)$ as the temporal axis by normalizing the temporal length within $[0,1]$. Specifically, for the $i$-th frame in a video of length $T$, the action variate is thus given by
\begin{eqnarray}
\label{Eq:a}
a(i)&=&\frac{i}{T}.
\end{eqnarray}
}

\begin{figure}[!t]
\begin{center}
\subfigure{
\includegraphics[width=0.9\columnwidth]{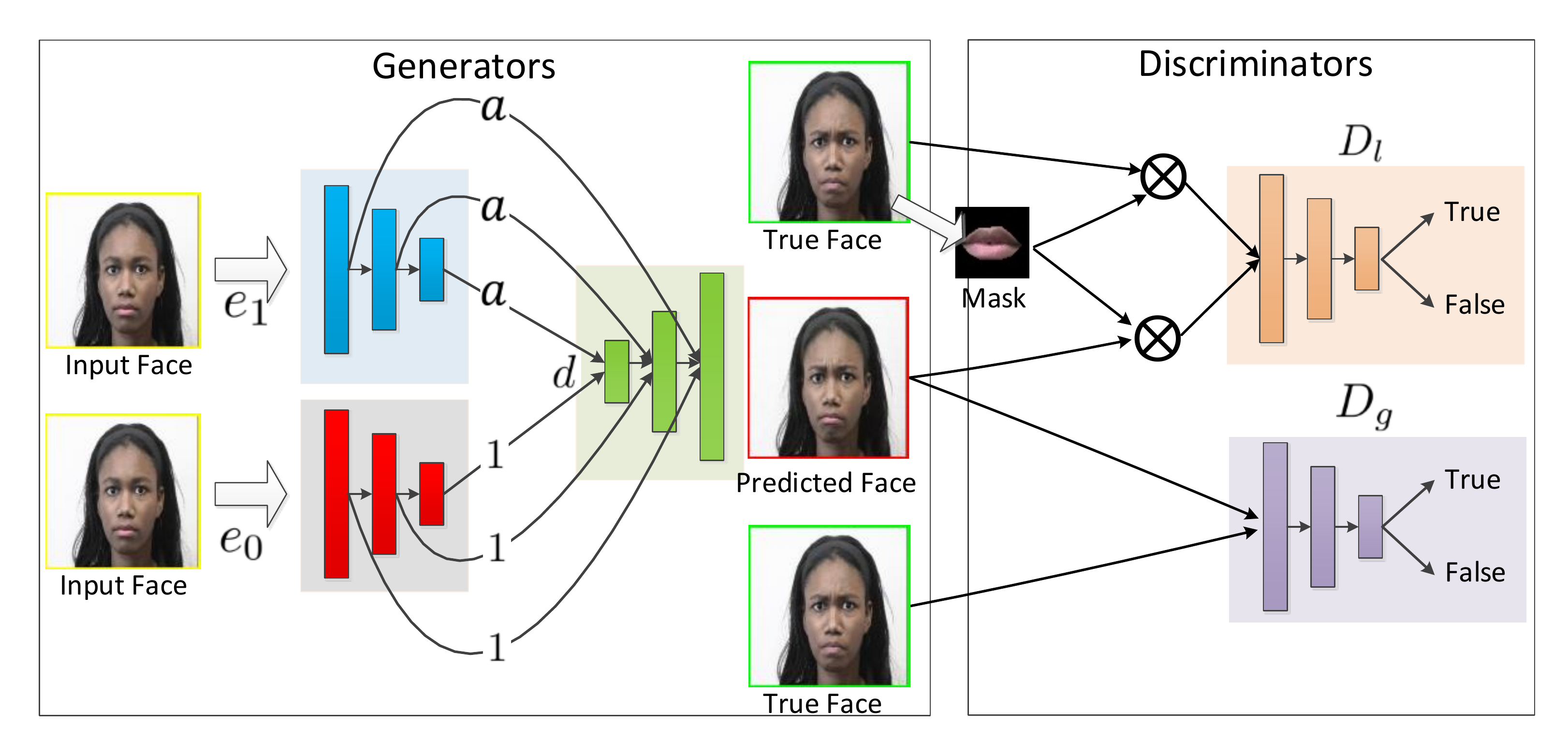}
}
\vskip -0.2in
\caption{Illustration of our model. It consists of two encoders $e_0$ and $e_1$, one decoder $d$, and two discriminators $D_l$ and $D_g$.}
\label{Fig:network}
\end{center}
\vspace{-15pt}
\end{figure}

\subsection{Network design for the video frame generator $f(\Mat{I},a)$}
\label{Sec:network}
Figure~\ref{Fig:network} sketches the neural network modules we designed for the video frame generator $f(\Mat{I},a)$. It is mainly composed of three modules: a base encoder, a residual encoder, and a decoder. In addition, there are two two discriminators for the purpose of generative adversarial training. We employ the Convolution-BatchNorm-ReLu layers in these modules~\cite{ioffe2015batch}.

\textbf{Generator.}
Considering that $f(\Mat{I},0)=I$,
a straightforward construction of $f$ is to linearly combine the input image with a residual term. However, it would incur severe artifacts to add the two in the pixel space. Instead, we perform linear aggregation in the feature space. Denote by $e_0(\cdot)$ and $e_1(\cdot)$ the base encoder and the residual encoder, respectively, where the former is to extract the feature hierarchy for self-reconstruction and the latter takes care of the change that is useful for constructing the future frames. Concretely, we have the following,
\begin{eqnarray}
\label{Eq:agg}
\Mat{F}(\Mat{I},a) &=& e_0(\Mat{I}) + a\cdot e_1(\Mat{I}),
\end{eqnarray}
where the variate $a$ explicitly determines the intensity of shift off the base encoder. Note that the summation in eq.~\ref{Eq:agg} is layer by layer (cf.\ Figure~\ref{Fig:network}).
The resulting feature hierarchy $\Mat{F}$ is then fed to the decoder $d$ for video frame generation, \ie,
\begin{eqnarray}
\label{Eq:dec}
\Mat{V}(a) :=f(\Mat{I},a)= d(\Mat{F}(\Mat{I},a))=d(e_0(\Mat{I})+a\cdot e_1(\Mat{I})),
\end{eqnarray}
where the decoder mirrors the base encoder's architecture and takes as input the feature hierarchy in the reverse order (cf.\ Figure~\ref{Fig:network}). 



\textbf{Discriminators.}
We use two discriminators for the purpose of adversarial training: a global discriminator $D_g$ and a local one $D_l$. The global discriminator contrasts the generated video frame $\Mat{V}(a)$ to the groundtruth frame. This is a standard and effective practice in video generation~\cite{vondrick2016generating,tulyakov2017mocogan} and image-to-image translation~\cite{isola2017image}. In addition, we employ a local discriminator to take special account of certain local parts of interest. Taking the smile expression for example, the mouth region is the most active part and deserves more detailed synthesis than the others. We first compute a mask as a convex closure of the detected facial landmarks around the subject's mouth and then filter out the mouth regions by the mask for both groundtruth and the generated frames. The local discriminator $D_l$ is then applied to the filtered pairs. 

\subsection{Training loss}
\label{Sec:training}
We prepare training data in the following manner. Given a video clip of length $T$, assume it has been labeled such that the 1st frame is in the neutral expression state and the $T$-th is at the peak of the expression. We assign coefficient $a=(t-1)/(T-1)$ to the $t$-th frame of this clip. Denote by $\Mat{Y}(a)$ one of these groundtruth frames. We train our neural network using the adversarial loss
\begin{eqnarray}
\label{Eq:gan}
\mathcal{L}_{g} &:=& -\log(1-D_g(\Mat{V}(a))) - \log D_g(\Mat{Y}(a))),
\end{eqnarray}
for the global discriminator $D_g$ and
\begin{eqnarray}
\label{Eq:gan}
\mathcal{L}_{l} &:=& -\log(1-D_l\left(\Mat{V}(a)\circ \Mat{M}(a)\right))- \log D_l\left(\Mat{Y}(a)\circ \Mat{M}(a)\right),
\end{eqnarray}
for the local discriminator $D_l$, where $\Mat{M}(a)$ is the  mask to crop out the local patch of interest and $\circ$ denotes element-wise multiplication. Joining the previous work~\cite{isola2017image}, we find that it is beneficial to augment the adversarial loss with the reconstruction error: $\mathcal{L}_{r} = \|\Mat{Y}(a)-\Mat{V}(a)\|_1$.

\textbf{Temporal continuity.}
The generative adversarial training of the neural network may result in mode collapse (different modes collapse to a mixed mode that does not exist in the real data) and mode dropping (the generator fails to capture some of the modes). Whereas the reconstruction loss $\mathcal{L}_r$ alleviates these issues to some extent, it is defined at a particular time step and does not track the temporal continuity in the video. We propose to regularize the difference between nearby video frames generated by the network. The regularization both helps prevent the mode dropping issue and makes the generated video clips smooth over time. It is defined as below,
\begin{eqnarray}
\mathcal{R}_{t} &:=& \|\Mat{V}(a)-\Mat{V}(a-\Delta a)\|_1 + \|\Mat{V}(a)-\Mat{V}(a+\Delta a)\|_1,
\end{eqnarray}
where $\Delta a$ is a small increment. The training is still frame-wise and efficient as the frames $\Mat{V}(a)$, $\Mat{V}(a+\Delta a)$, and $\Mat{V}(a-\Delta a)$ are computed independently from each other.

\eat{
For this purpose, we each time compare the difference between $\Mat{V}(t)$ and its simulated neighbourhoods $\Mat{V}(t-\Delta t)$ and $\Mat{V}(t+\Delta t)$, where $\Mat{V}(t-\Delta t) = f(\Mat{I}_0, a(t)-\Delta a)$, $\Mat{V}(t+\Delta t) = f(\Mat{I}_0, a(t)+\Delta a)$, and $\Delta a$ is a small increment. The training is still frame-wise and thus efficient as computing $\Mat{V}(t)$ and its neighbourhoods are independent to each other.
}

\textbf{Facial landmark prediction.}
As discussed earlier, we use facial landmarks to extract the local regions of interest for the local discriminator. In our experiments, we use the Dlib Library~\cite{dlib09} to detect 68 landmarks from any groundtruth video frame. These landmarks are supposed to be at the same locations for the correspondingly synthesized video frame. Therefore, we stack another 68-dimensional channel on the top of the second-to-last layer of the global decoder to predict the landmarks, enforcing the generator to provide details of the face. This loss is denoted by $\mathcal{L}_{k} := \|\Mat{\bar{K}}(a)-\Mat{K}(a)\|_2$, where $\Mat{\bar{K}}$ and $\Mat{K}(t)$ are the predicted and groundtruth landmarks, respectively.

Putting the above together, we train our neural networks by alternating between optimizing the generator and the discriminators in order to solve the following problem,
\begin{align}
\min_{D_g,D_l}\max_{f} \;\sum_a \mathcal{L}_g+\mathcal{L}_l+\mathcal{L}_k + \mathcal{R}_t.
\end{align}
In the experiments, we use slightly different weights in front of the loss and regularization terms.

\eat{
As discussed in last section, we also have a regulation loss to penalize the temporal continuity.  Hence, we minimize the difference between $\Mat{V}(t)$ and its simulated neighbourhoods $\Mat{V}(t-\Delta t)$ and $\Mat{V}(t+\Delta t)$ via the $\ell_1$ distance, \ie,
\begin{eqnarray}
\label{Eq:lt}
\mathcal{L}_{t} &=& \|\Mat{V}(t)-\Mat{V}(t-\Delta t)\|_1 + \|\Mat{V}(t)-\Mat{V}(t+\Delta t)\|_1.
\end{eqnarray}

To enable our network for landmark prediction, we additionally formulate the $\ell_2$ loss as
\begin{eqnarray}
\label{Eq:lland}
\mathcal{L}_{land} &=& \|\Mat{\bar{K}}(t)-\Mat{K}(t)\|_2,
\end{eqnarray}
where $\Mat{\bar{K}}$ and $\Mat{K}(t)$ are the predicted and ground-truth landmarks, respectively.

Putting the losses all together, we train the generation components (\ie $e_0$, $e_1$ and $d$) by
\begin{eqnarray}
\label{Eq:l-gen}
\mathcal{L}_{G} &=& \mathcal{L}_{gen} + \alpha \mathcal{L}_{gen}^{(m)} + \beta \mathcal{L}_{\ell_1} + \gamma \mathcal{L}_{t} + \lambda \mathcal{L}_{land},
\end{eqnarray}
and the discriminators (\ie $D_g$ and $D_l$) by
\begin{eqnarray}
\label{Eq:l-dis}
\mathcal{L}_{D} &=& \mathcal{L}_{dis} + \alpha \mathcal{L}_{dis}^{(m)},
\end{eqnarray}
where $\alpha$, $\beta$, $\gamma$ and $\lambda$ are constant parameters. As we employ the Stochastic Gradient Decent (SGD) for the training, the losses in Eq.~\eqref{Eq:l-gen} and Eq.~\eqref{Eq:l-dis} need to be summed over all mini-batch videos.
}

\subsection{Jointly learning the models of different types of facial expressions}
\label{Sec:multi-label}
Thus far, we have assumed a separate model $f(\Mat{I},a)$ for each type of facial expressions. It is straightforward to extend it to handle $n>1$ types of emotions jointly:
\begin{align}
\label{Eq:mult}
\Mat{V}(t) = d(\Mat{F}(\Mat{I}, \Vec{a}(t))), \quad \Mat{F}(\Mat{I},\Vec{a}(t)) = e_0(\Mat{I}) + \sum\nolimits_{i=1}^n a_i(t)\cdot e_i(\Mat{I}), \quad \Vec{a}(t)\in[0,1]^{n}
\end{align}
where $\Vec{a}(t)$ is an $n$-dimensional vector with each dimension standing for one emotion type. Since each training video clip contains one type of emotion, only one entry of the vector $\Vec{a}(t)$ is non-zero in the training stage. At the test stage, however, we examine the effect of mixing some emotions by allowing non-zeros values in multiple entries of the vector $\Vec{a}(t)$. Note that different types of emotions share the same base encoder $e_0$ (as well as the decoder $d$ and discriminators $D_g, D_l$) and differ only by the residual encoders $e_1,e_2,\cdots,e_n$.

\eat{
The network with the multiple-action variates shares the similar structure as the single-action model, excluding that we apply $n$ residual encoders for feature aggregation as
\begin{eqnarray}
\label{Eq:agg-mul}
\Mat{F}(t) &=& e_0(\Mat{I}_0) + \sum\nolimits_{i=1}^n a_i(t)\cdot e_i(\Mat{I}_0).
\end{eqnarray}
The aggregated feature is then proceeded by the decoder to predict the video frame.

The multiple-action model enables us to generate videos for different actions simultaneously, which is more efficient than the separate training by adopting different models to different actions. More importantly, learning multiple actions with one single model is able to discover the correlations between different actions, which helps us to transfer the action from one domain to the other. We provide the experimental evaluations on multiple-action models in \textsection~\ref{Sec:trans}.
}

\section{Experiments}
Given a neutral face image and a target expression(\eg, smile), we generate a video clip to simulate how the face will change towards the target expression. We not only use the public CK+~\cite{lucey2010extended} dataset for model training  but also significantly extend it in scale. The new larger-scale dataset is named CK++. To better evaluate the performance of our method, we further collect around raw $150$ face images from the Web. We then generate the facial expression videos based on these collected photos and submit them to the AMT for rating.

\textbf{CK+.}
The Extended Cohn-Kanade (CK+) dataset~\cite{lucey2010extended} is a widely used dataset for facial emotion analysis. It contains 593 videos of  8 different emotion categories (including the neutral category) and 123 subjects. Each video frame is provided with a 68-point facial landmark label. We use three major categories (\ie, ``happy'', ``angry'', and ``surprised'') in this paper.

\textbf{CK++.}
Most images in CK+ are in the gray-scale. We augment CK+ by additionally collecting the facial expression videos in the RGB-scale.
The videos are collected by a fixed camera from 65 volunteers consisting of 32 males and 33 females.
Each volunteer is asked to perform each of the  ``happy'', ``angry'' and ``surprised'' expressions for at least twice.
We manually remove the redundant frames before the initial neutral state and after the stationary peak state. We also remove the videos that contain severe head movement or blurry faces. There are  214, 167, and 177 video clips for the ``happy'', ``angry'', and ``surprised'' expressions, respectively. On average, each clip has 21 frames. Finally, we use the Dlib Library~\cite{dlib09} to detect 68 landmarks from each of the frames.

\subsection{Implementation Details}
For our encoders, we employ eight-downsampling-layer architectures with the Leaky-ReLu activation function. The decoder mirrors the encoder's architecture by eight upsampling layers and yet the ReLu activation function. 
Inspired by the U-Net~\cite{ronneberger2015u}, we further add skip connections between intermediate layers of the encoders and the decoder (cf.\ Figure~\ref{Fig:network}).
Both the global and local discriminators are constructed by concatenating 3 convolution layers.

We use 10 video clips from the CK++ dataset for validation and all the others for training.
Our network is trained from scratch with all parameters normally initialized. For each training batch, we randomly sample a video clip and then use its first frame as the input image to train the network. All images are resized to 289x289 and randomly cropped to 256x256 before being fed into the network.
The Adam optimizer is used in the experiments, with the initial learning rate of 0.0002. The whole training process takes 2100 epochs, where one epoch means a complete pass over the training data.
As discussed in \textsection~\ref{Sec:network}, training the local discriminator requires a mask to crop the local regions of interest. Since mouth is the most expressive region, we crop it out by a convex closure of the landmarks around the mouth. We set the small increment to $\Delta a=0.1$ for temporal regulation $\mathcal{R}_t$. 


As our task of controllable image-to-video translation is new, there is no exactly related method in the literature.
Conservatively, we adapt two previous methods to our experiments including Hierarchical Prediction (HP)~\cite{villegas2017learning} and Convolution-LSTM (ConvLSTM)~\cite{xingjian2015convolutional}. In particular, we make the following changes to HP and ConvLSTM to fit them to our problem: 1)
Since both HP and CovLSTM use LSTM to recursively generate video frames, we have to fix the length of the video sequence to be generated. We do so by uniformly sampling 10 frames per video clip. 2) We train a separate model for each target expression. 3) We replace the CovNet in ConvLSTM and the Visual-Structure in HP with U-Net because the results of their default architectures works not well.


\begin{figure}[!t]
\begin{center}
\subfigure{
\includegraphics[width=\columnwidth]{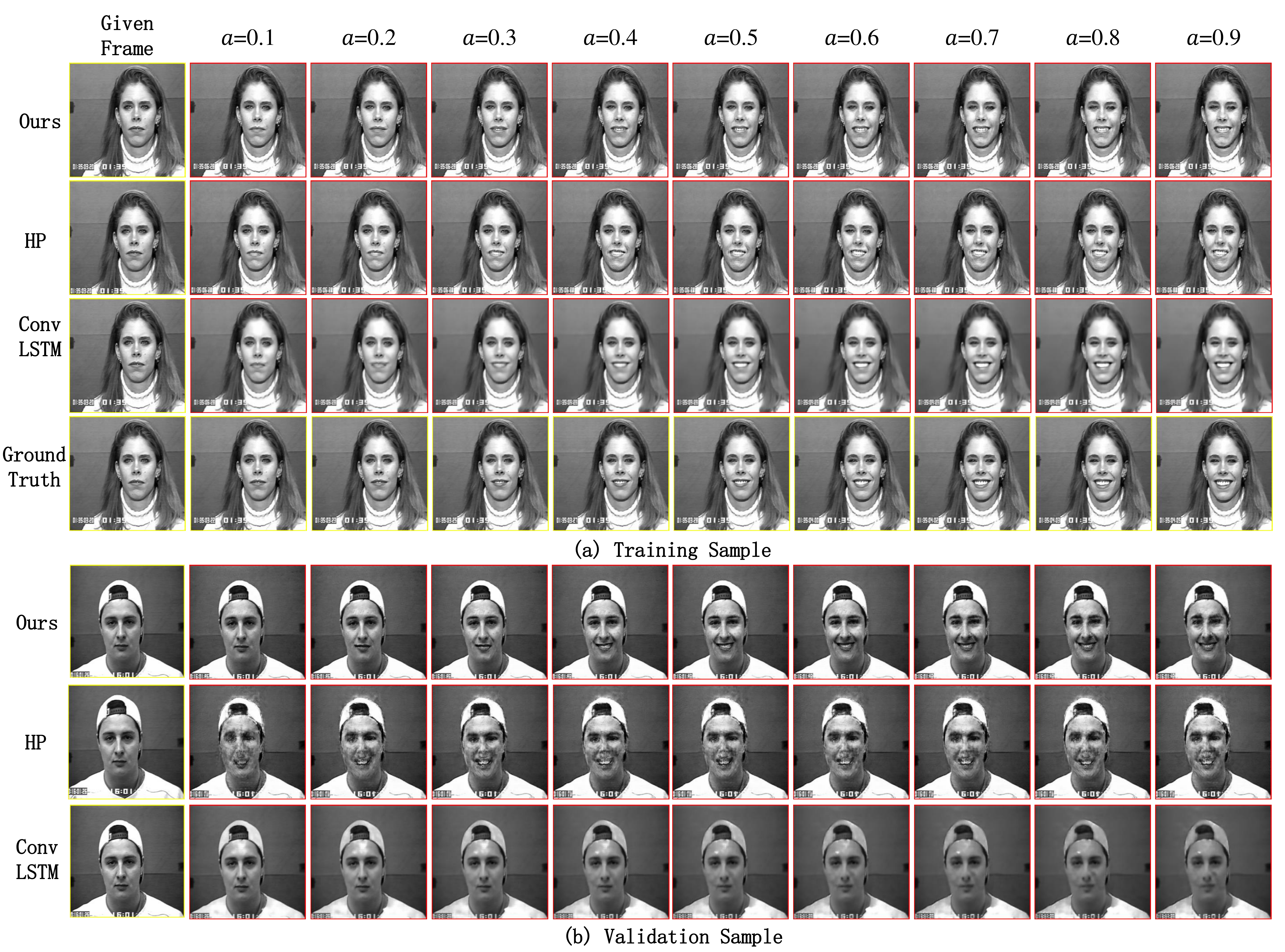}
}
\vskip -0.2in
\caption{Visualization for the ``happy'' expression by different methods.}
\label{Fig:visualization}
\end{center}
\vspace{-23pt}
\end{figure}

\begin{figure*}[h]
\vskip -0.15in
	\centering{
    \includegraphics[width=5cm]{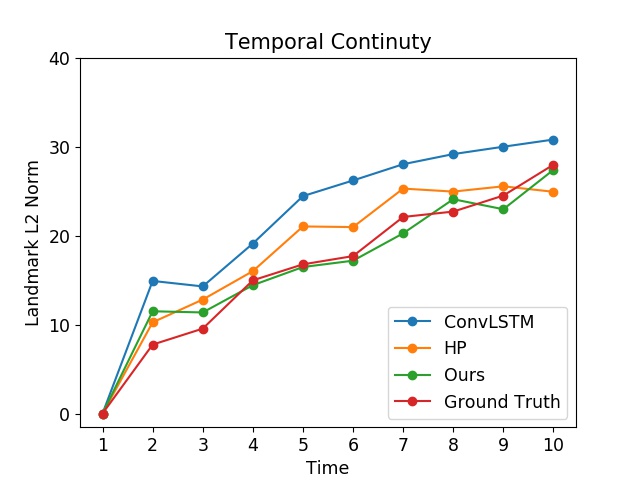}
\includegraphics[width=5cm]{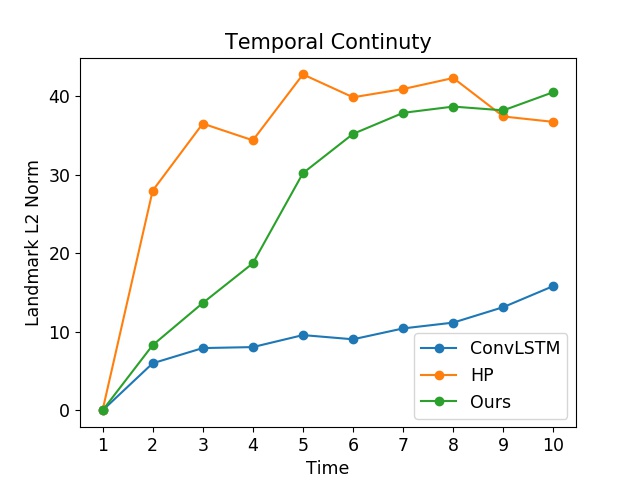}
	}
\vskip -0.1in
	\caption{The L2 norm between the landmarks in each frame and the initial frame over time. The left and right sub-figures are on the training and validation samples, respectively.}
	\label{fig:temporal_fig}
    \vspace{-12pt}
\end{figure*}

\subsection{Evaluations}

\textbf{Visualization.}
For a fair comparison, we let our method output the same number of frames as HP and ConvLSTM do (e.g., 10) by setting $a=\{0.1, 0.2,\cdots, 1\}$. Figure~\ref{Fig:visualization} displays the generated video frames of the ``happy'' emotion for two persons, one seen at training and the other unseen. We can see that both our method and the LSTM-based baseline models perform well on the training image. However, when it comes to the person of the validation set, our model clearly outperforms the baselines in terms of both image quality and temporal continuity. 

\textbf{Analysis on temporal continuity.}
To evaluate temporal continuity quantitatively, we extract facial landmarks from each generated video frame (\emph{i.e.} a 68x2 dimensional vector) and then compute the L2 distance between the landmarks of each frame and those of the initial one. Figure~\ref{fig:temporal_fig} plots the distances versus the time steps on the same training and validation samples in Figure~\ref{Fig:visualization}.
We find that the curve of our approach aligns well with that of the groundtruth video frames.
the face in the image sequence generated by ConvLSTM doesn't seem to move much, as the facial keypoints almost remain in the same place as the generation process goes.
As for HP baseline, there is a sudden change between the generate frame and the initial frame, which means the expression generate by HP do not have a good temporal continuity. In fact, in our validation data, we see many cases  where the person in the generated video by HP seem to smile a little, go back, and then smile again. We expect to avoid this phenomenon in our task.
While for our proposed model, the L2 norm between keypoints in each frame and the initial frame grows steadily and linearly without decreasing, showing the images generated by our method have a good temporal continuity.

\textbf{AMT results.}
Following~\cite{villegas2017learning}, we also conduct user studies to compare the results of different methods.
For this purpose, we formulate test data by downloading about $150$ face images from the Web without any post-processing. We then generate the facial expression videos of ``happy'', ``surprise'', and ``angry'' by our approach and two baselines.
We pair the video clip of the same input image for the same emotion by our method with that by either of the baselines, and then ask an AMT worker to choose which one is more realistic in terms of the temporal continuity, image quality, naturalness of the expressions, etc.  The rows above the last in Table~\ref{Tab:compare} show that the users prefer our results to either of the baselines' to a large margin. We also ask users to choose the most realistic clip from three, respectively generated by our model and the two baselines. As shown in the last row of Table~\ref{Tab:compare}, our results are again selected significantly more often than the other two.

In addition, we perform a more challenging test by mixing the simulated videos by our method with  real videos, and then asking an AMT worker to judge if the displayed video is real or not. As reported in Table~\ref{Tab:real}, the results are encouraging, as nearly 50\% of our generated videos from the test faces are labeled as real by AMT workers.

\begin{figure}[!t]
\vskip -0.2in
\begin{center}
\subfigure{
\includegraphics[width=0.9\columnwidth]{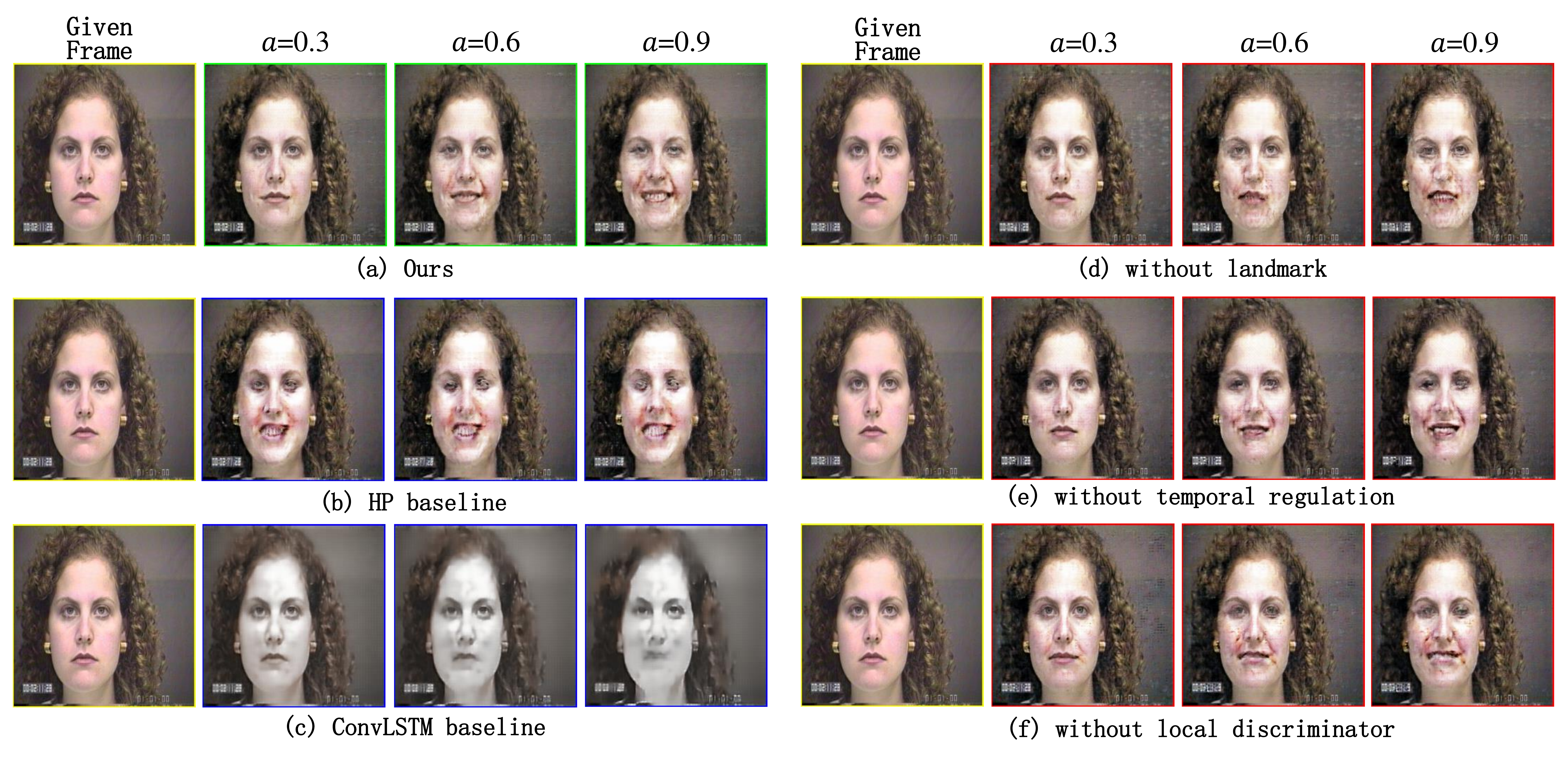}
}
\vskip -0.2in
\caption{Ablation studies on our method.}
\label{Fig:compare}
\end{center}
\vspace{-15pt}
\end{figure}

\begin{figure}[!t]
\begin{center}
\subfigure{
\includegraphics[width=0.9\columnwidth]{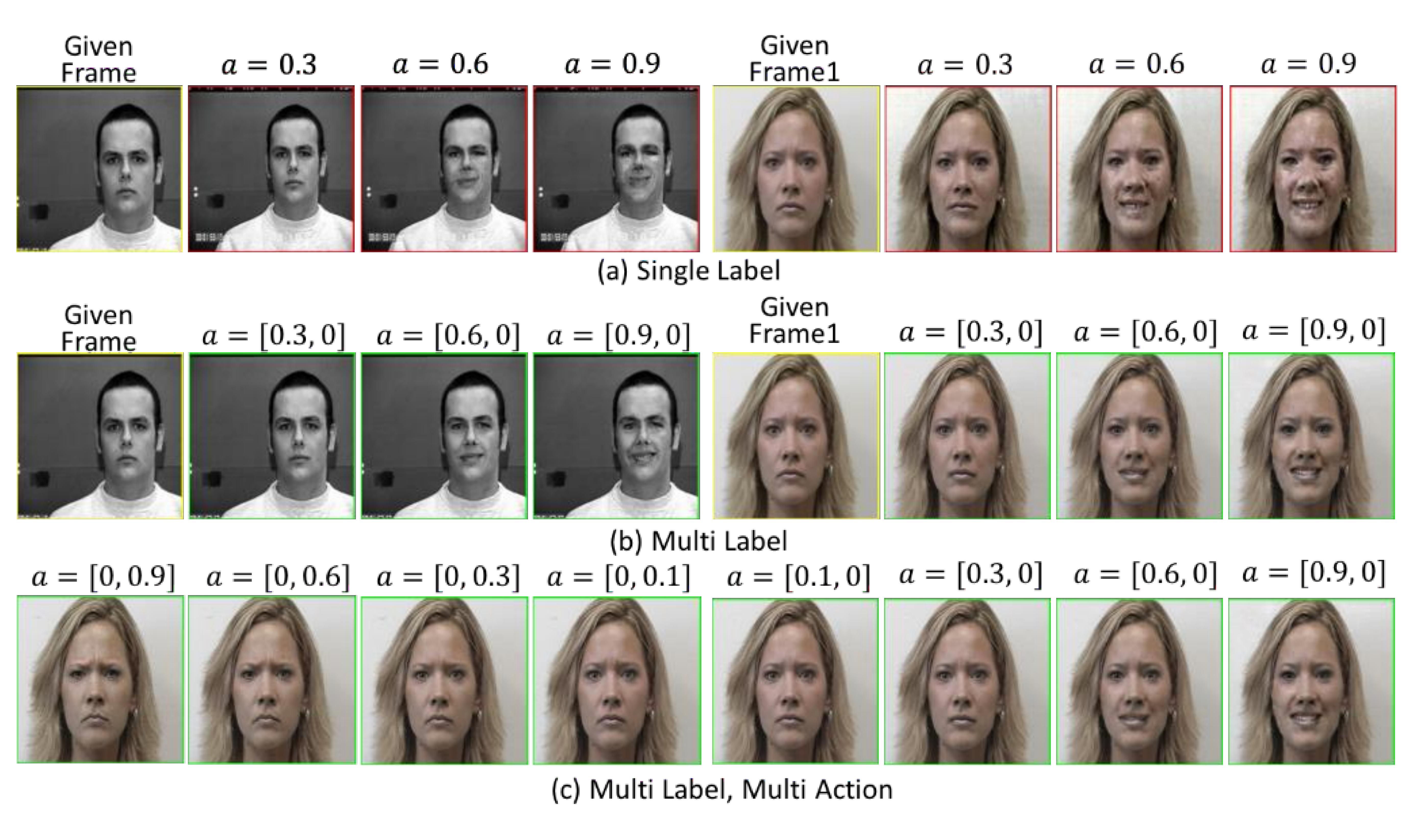}
}
\vskip -0.2in
\caption{Evaluation on multiple-label models. (a) Videos generated by single-label models; (b) Videos generated by multi-label models; (c) Transferring the ``angry'' expression to the ``happy'' one by controlling the action variate in the multi-label model. }
\label{Fig:transfer}
\end{center}
\vspace{-25pt}
\end{figure}

\begin{table}[t!]
\centering
\caption{Comparisons of AMT results between our method against two baselines.}
\label{Tab:compare}
\tabcolsep 8pt 
\renewcommand{\arraystretch}{0.8}
\begin{tabular}{l|ccc|c}
\toprule
"Which video looks more realistic?"   & Happy        & Surprise        & Angry           & Mean         \\
\midrule
Prefers ours over ConvLSTM         & 83.8\%         & 82.2\%           & 83.0\%           & 83.0\%            \\
Prefers ours over HP               & 77.3\%          & 69.7\%           & 67.1\%           & 71.4\%        \\
Prefers ours over both baselines   &  69.9\%         & 63.7 \%           & 61.8\%           & 65.1\%        \\
\bottomrule
\end{tabular}
\vskip -0.2in
\end{table}

\begin{table}[ht!]
\centering
\vskip -0.1in
\caption{AMT results on how many videos generated by our method can fool the workers.}
\label{Tab:real}
\tabcolsep 8pt 
\renewcommand{\arraystretch}{0.8}
\begin{tabular}{l|ccc|c}
\toprule
"Is this video real?"   & Happy        & Surprise        & Angry           & Mean         \\
\midrule
Training Videos     &  64.5\%   &  59.7\%         &  57.3\%    & 60.5\%  \\
Testing Videos      &  49.4\%   &  52.2\%         &  48.3\%    & 49.9\%   \\
\bottomrule
\end{tabular}
\vspace{-10pt}
\end{table}

\subsection{Other analysis}
\textbf{Ablation Studies.}
We have run some ablation studies to examine some key components of our approach, as illustrated in Figure~\ref{Fig:compare}. We implement several variants of our method without the local discriminator, without predicting the landmarks, and without the temporal continuity regularization.
\textbf{I.} People easily focus on mouths when they first see a video. So a local discriminator on the mouth would make the video seem more realistic to audience. Without local discriminator, Figure~\ref{Fig:compare} (f) easily involves blurring artifact compared to the original model.
\textbf{II.}
Landmark prediction gives a higher level regulation, which can enable our model to have the ability to generate facial feature in the right place, avoiding generating multiple features in the same image, therefore avoid blurring artifact and make the generate image more clear and reasonable.
\textbf{III.}
We can see from the example in Figure~\ref{Fig:compare} (e), temporal regulation not only force the movement perform continually, avoid sudden change, but also have the effect to make the generated image more clear.

\textbf{Controlling Action Variable.}
One of the most interesting part in our proposed approach is that we can control the lengths of the videos by the action variable $a$. We provide demos on controlling the action variable in the supplementary materials.

\textbf{Jointly model multiple types of expressions.}
We present in \textsection~\ref{Sec:multi-label} that our model is applicable for learning different types of emotions simultaneously. As a result, we may mix different emotions by providing more than one non-zero entries to the vector $\Vec{a}(t)$. We first show that it gives rise to better results to simultaneously model different types of expressions in one neural network than learning a separate model for each. To show this, we further formulate a training set by using the faces of the ``happy'' and ``angry'' emotions. Each training person has either emotion but not both. For example, the two persons in Figure~\ref{Fig:transfer} have only the ``angry'' expression. Figure~\ref{Fig:transfer} (a) and (b) demonstrate that the generated ``happy'' videos by the jointly modeling the two emotions is more realistic than the models of individual emotions. We conjecture that it is due to the strong correlation between emotions that enables information sharing between the residual encoders.

Another interesting application for the joint model is that it can easily perform transfer between two different emotions by using proper values of the action variable, as illustrated in Figure~\ref{Fig:transfer} (c). More results are included in the supplementary materials.
\vspace{-10pt}

\section{Conclusion}
In this paper, we study image-to-video translation with a special focus on the facial expression videos. We propose a user-controllable approach so as to generate video clips of various lengths and different target expressions from a single face image. Both the lengths and types of the expressions can be controlled by users. To this end, we design a novel neural network architecture that can incorporate the user input and also propose several improvements to the adversarial training method for the neural networks. Experiments and user studies verify the effectiveness of our approach. It would be interesting to investigate the image-to-video translation in domains other than the facial expressions in the future work. In addition, we will explore the potential of progressive training~\cite{karras2017progressive} for generating higher-definition video clips from a single input image.




\bibliographystyle{unsrt}
\small
\bibliography{ref}

\end{document}